\documentclass[onecolumn]{article} 
\usepackage{arxiv}
\usepackage[square,sort,comma,numbers]{natbib}
\usepackage[utf8]{inputenc} 
\usepackage[T1]{fontenc}    
\usepackage[shortlabels]{enumitem}
\usepackage[obeyspaces]{url}
\usepackage{wrapfig}
\usepackage{booktabs}       
\usepackage{amsfonts}       
\usepackage{nicefrac}       
\usepackage{microtype}      
\usepackage{graphicx}
\usepackage{natbib}
\usepackage{doi}
\usepackage{multicol}
\usepackage{multirow}
\usepackage{tabularx}
\usepackage{tabulary}
\usepackage{makecell}
\usepackage{tabularx,ragged2e}
\usepackage{fancyhdr, blindtext}
\usepackage{subfigure}
\usepackage{float}
\usepackage{hyperref}
\usepackage{hhline}
\usepackage{array}
\usepackage{amsmath}
\usepackage{diagbox}
\usepackage{color}
\usepackage{textcomp}
\usepackage{comment}
\usepackage{relsize}
\usepackage[symbol]{footmisc}
\usepackage{caption}
\usepackage{chngcntr}
\usepackage{amssymb}%
\usepackage{amsthm}%
\usepackage{mathrsfs}%
\usepackage[title]{appendix}%
\usepackage{xcolor}%
\usepackage{textcomp}%
\usepackage{manyfoot}%
\usepackage{booktabs}%
\usepackage{algorithm}%
\usepackage{algorithmicx}%
\usepackage{algpseudocode}%
\usepackage{listings}%
\usepackage{natbib} 
\usepackage{comment}
\usepackage{etex} 
\usepackage{morewrites}
\usepackage{authblk} 
\usepackage[capitalise]{cleveref}

\renewcommand{\headeright}{AI for Surgery Safety}
\renewcommand{\shorttitle}{BetaMixer: Intraoperative Adverse Events Detection}
\renewcommand{\undertitle}{Artificial Intelligence for Surgical Safety}

\setcitestyle{square}
\hypersetup{
pdftitle={BetaMixer},
pdfsubject={q-bio.NC, q-bio.QM},
pdfauthor={Bose et al.},
pdfkeywords={classification, regression, anomaly detection, bleeding segmentation},
}

\title{Feature Mixing Approach for Detecting Intraoperative Adverse Events in Laparoscopic Roux-en-Y Gastric Bypass Surgery}

\date{}

\author{ 
    \textbf{Rupak Bose}\textsuperscript{1}, 
    \textbf{Chinedu Innocent Nwoye} \textsuperscript{1,4,*},
    \textbf{Jorge Lazo}\textsuperscript{1},
    \textbf{Joël Lukas Lavanchy} \textsuperscript{3,4,†},
    \textbf{Nicolas Padoy}\textsuperscript{1,4,†}\\
    \textsuperscript{1}ICube, UMR7357, CNRS, INSERM, University of Strasbourg, France\\    
    \textsuperscript{2}{University Digestive Health Care Center, Clarunis, Basel, Switzerland}\\
    \textsuperscript{3}{Dept. of Biomedical Engineering, University of Basel, Switzerland}\\
    \textsuperscript{4}IHU Strasbourg, France\\
    {\smaller        
        \textsuperscript{*}Corresponding author: nwoye@unistra.fr,~ \textsuperscript{†}{Co-last authors}.\\
    }
}

\begin{document}
\maketitle



\begin{abstract}
    Intraoperative adverse events (IAEs), such as bleeding or thermal injury, can lead to severe postoperative complications if undetected. However, their rarity results in highly imbalanced datasets, posing challenges for AI-based detection and severity quantification. We propose BetaMixer, a novel deep learning model that addresses these challenges through a Beta distribution-based mixing approach, converting discrete IAE severity scores into continuous values for precise severity regression (0-5 scale). BetaMixer employs Beta distribution-based sampling to enhance underrepresented classes and regularizes intermediate embeddings to maintain a structured feature space. A generative approach aligns the feature space with sampled IAE severity, enabling robust classification and severity regression via a transformer. Evaluated on the MultiBypass140 dataset, which we extended with IAE labels, BetaMixer achieves a weighted F1 score of 0.76, recall of 0.81, PPV of 0.73, and NPV of 0.84, demonstrating strong performance on imbalanced data. By integrating Beta distribution-based sampling, feature mixing, and generative modeling, BetaMixer offers a robust solution for IAE detection and quantification in clinical settings.
    
\end{abstract}

\begin{figure}[ht]
  \begin{center}
    \includegraphics[width=0.8\textwidth,height=2.3in]{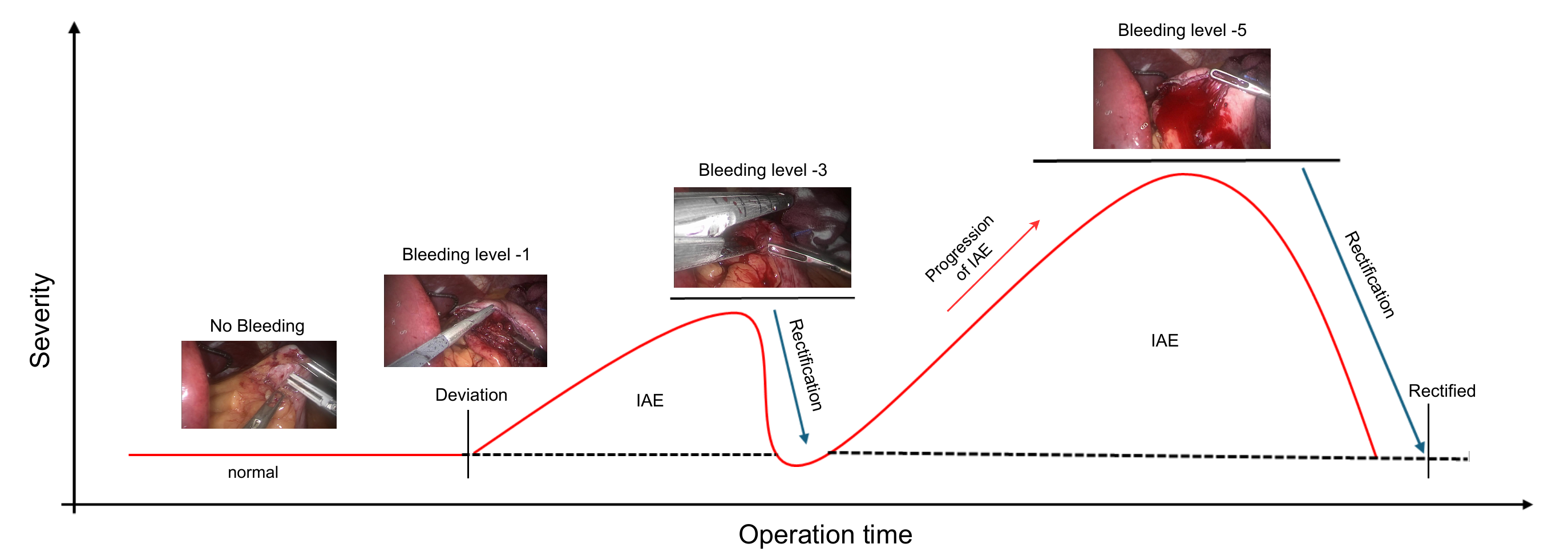}
    \caption{Illustration of an occurrence of intraoperative adverse event---a case of bleeding---showing different levels of severity.}
    \label{fig:qualitative_visualization}
  \end{center}
\end{figure}



\enlargethispage{\baselineskip}


\twocolumn


\section{Introduction}
\label{sec:intro}

Intraoperative adverse events (IAEs), such as bleeding, thermal injury, and mechanical injury, are rare but critical occurrences during surgery that can lead to severe postoperative complications, including infection, organ dysfunction, or even mortality. These events not only jeopardize patient safety but also increase healthcare costs due to prolonged recovery times. Timely detection and accurate IAE severity quantification are crucial for enabling prompt intervention and improving surgical outcomes. However, traditional manual monitoring by surgical teams is prone to human error, underscoring the need for automated, real-time detection systems \cite{mitchell2016patient}.

Artificial Intelligence (AI) has emerged as a promising tool for automating IAE recognition and quantification, offering real-time feedback to surgeons \cite{eppler2023automated}. Despite this potential, the rarity of IAEs results in highly imbalanced datasets, which pose significant challenges for training effective AI models. Standard detection techniques often struggle with such imbalances, and the inherent complexity of surgical procedures further complicates the identification of deviations from normal workflow. Additionally, quantifying the severity of IAEs—ranging from mild to critical—is essential for determining the appropriate surgical response, yet this aspect remains underexplored in existing literature.
Current approaches to surgical anomaly detection often fail to address the diversity of IAEs or the importance of severity quantification, limiting their practical utility in clinical settings \cite{beyersdorffer2021detection}. 

To bridge this gap, we propose \textit{BetaMixer}, a deep learning-based framework designed for both IAE classification and their severity regression during Roux-en-Y gastric bypass surgery. BetaMixer leverages a Beta ($\beta$) distribution to transform discrete IAE labels into continuous variables, enabling precise severity regression on a 0 to 5 scale. To address class imbalance, the model employs $\beta$ distribution-based sampling and regularizes intermediate embeddings to maintain a structured feature space. A generative component aligns the feature space with continuous severity labels, while a transformer network classifies IAEs and regresses their severity using a mean squared error (MSE) loss on sampled predictions and ground truths.

We evaluate BetaMixer on the MultiBypass140 dataset \cite{lavanchy2024challenges}, extended with IAE annotations. Our model achieves state-of-the-art performance, particularly for rare IAEs, with a weighted F1 score of 0.76, recall of 0.81, PPV of 0.73, and NPV of 0.84. These results highlight the importance of temporal context and continuous feature space modeling for accurate IAE detection and regression.

Our contribution is fourfold: 
(1) 
A unified framework for IAE classification and severity regression in Roux-en-Y gastric bypass surgery. 
(2) 
A Beta distribution-based method to model continuous severity from discrete annotations, addressing class imbalance via MSE loss. 
(3) 
A generative component to normalize feature space distributions, aligning with severity labels. 
(4) Superior results over baselines, with temporal models outperforming frame-based approaches, significantly improving IAE detection and regression across metrics.

\section{Related Work}
\label{sec:literature}
Detecting and mitigating Intraoperative Adverse Events (IAEs) is vital for improving surgical outcomes~\cite{eppler2023automated}. IAEs, arising from factors like human error, equipment malfunction, and patient responses, are rare in large video datasets, making their detection challenging. They are often annotated with class labels~\cite{wei2021intraoperative,hua2022automatic}, with their scale of severity often overlooked. The discrete nature of these annotations doesn't capture the continuous development of IAEs. Studies in other fields suggest that using distributions like the Beta distribution better models severity uncertainty~\cite{mariooryad2015cost}.

For their task proximity, anomaly detection methods, including supervised~\cite{kawachi2018complementary,checcucci2023development}, semi-supervised~\cite{Ruff2020Deep}, and unsupervised~\cite{audibert2020usad} approaches, are employed for IAE detection. Traditional models combine CNNs and RNNs for spatial and temporal tasks, while Transformers~\cite{dosovitskiy2020image} have recently outperformed them in capturing long-range dependencies. However, most existing work focuses on event classification~\cite{wei2021intraoperative,gawria2022classification}, leaving the quantification of IAE severity, especially for conditions like intraoperative bleeding and injuries, largely unexplored.


\section{Methods}
\label{sec:methods}

This section presents the methodology for developing \textit{BetaMixer}, a deep learning model designed for the classification and regression of the severity of intraoperative adverse events (IAEs) in Roux-en-Y gastric bypass surgeries. The goal is to enhance timely surgical intervention and improve clinical outcomes by accurately detecting IAEs and assessing their severity.

\subsection{Problem Definition}
Given a surgical video \( V = \{f_1, f_2, \dots, f_n\} \) consisting of \( n \) sequential frames, the task is to predict the IAE class \( \mathcal{C} \) and severity \( \mathcal{S} \) for each frame \( f_i \in V \). Here, \( \mathcal{C} \in \{\text{BL}, \text{MI}, \text{TI}\} \) represents the IAE categories (bleeding, mechanical injury, and thermal injury), and \( \mathcal{S} \in \{0, 1, \dots, m\} \) denotes the severity level, where \( m \) varies by IAE type (e.g., \( m=5 \) for bleeding).
To predict \( (\mathcal{C}_i, \mathcal{S}_i) \) for frame \( f_i \), we utilize a sequence of contiguous frames from the past \( k \) time steps, i.e., \( X_i = \{f_{i-k+1}, \dots, f_i\} \). The model learns a function \( \mathcal{F}(X_i) \rightarrow (\mathcal{C}_i, \mathcal{S}_i) \) that maps the current frame \( f_i \) to its IAE class and severity, leveraging temporal information from the preceding \( k \) frames.

\begin{table*}[!t]
    \centering
    \caption{Clinical definition of IAE severity scores.}
    \label{tab:def}
    \setlength\tabcolsep{0.09in}
    \resizebox{0.999\linewidth}{!}{%
    \begin{tabular}{@{}llll@{}}
        \toprule
        Severity &
        Bleeding &
        Thermal injury &
        Mechanical injury \\
        \midrule
        
        1 & \parbox{5cm}{Very low amount of blood lost} &
        \parbox{7cm}{Superficial penetration to “less vital” tissue} &
        \parbox{6cm}{Superficial penetration to "less vital" tissue, needle poke to tissue} \\
        
        2 & \parbox{5cm}{Low amount of blood lost} &
        \parbox{7cm}{Deep penetration to “less vital” tissue or any organ/tissue subjected to planned resection} &
        \parbox{6cm}{Full-thickness injury} \\
        
        3 & \parbox{5cm}{Intermediate amount of blood lost} & 
        \parbox{7cm}{Superficial penetration to “vital” tissue} &
        \parbox{6cm}{Superficial penetration to "vital" tissue} \\
        
        4 & \parbox{5cm}{High amount of blood lost} &
        \parbox{7cm}{Deep penetration to “vital” tissue to the level of muscularis/parenchyma} &
        \parbox{6cm}{Deep penetration to “vital” tissue} \\
        
        5 & \parbox{5cm}{Very high amount of blood lost} &
        \parbox{7cm}{Through and through injury to hollow organ or deeper parenchymal injury to solid organ} &
        \parbox{6cm}{Through and through injury to "vital" tissue}\\
        \bottomrule
    \end{tabular}
    }
\end{table*}

\subsection{Dataset}
\begin{figure}
    \centering
    \includegraphics[width=0.95\linewidth]{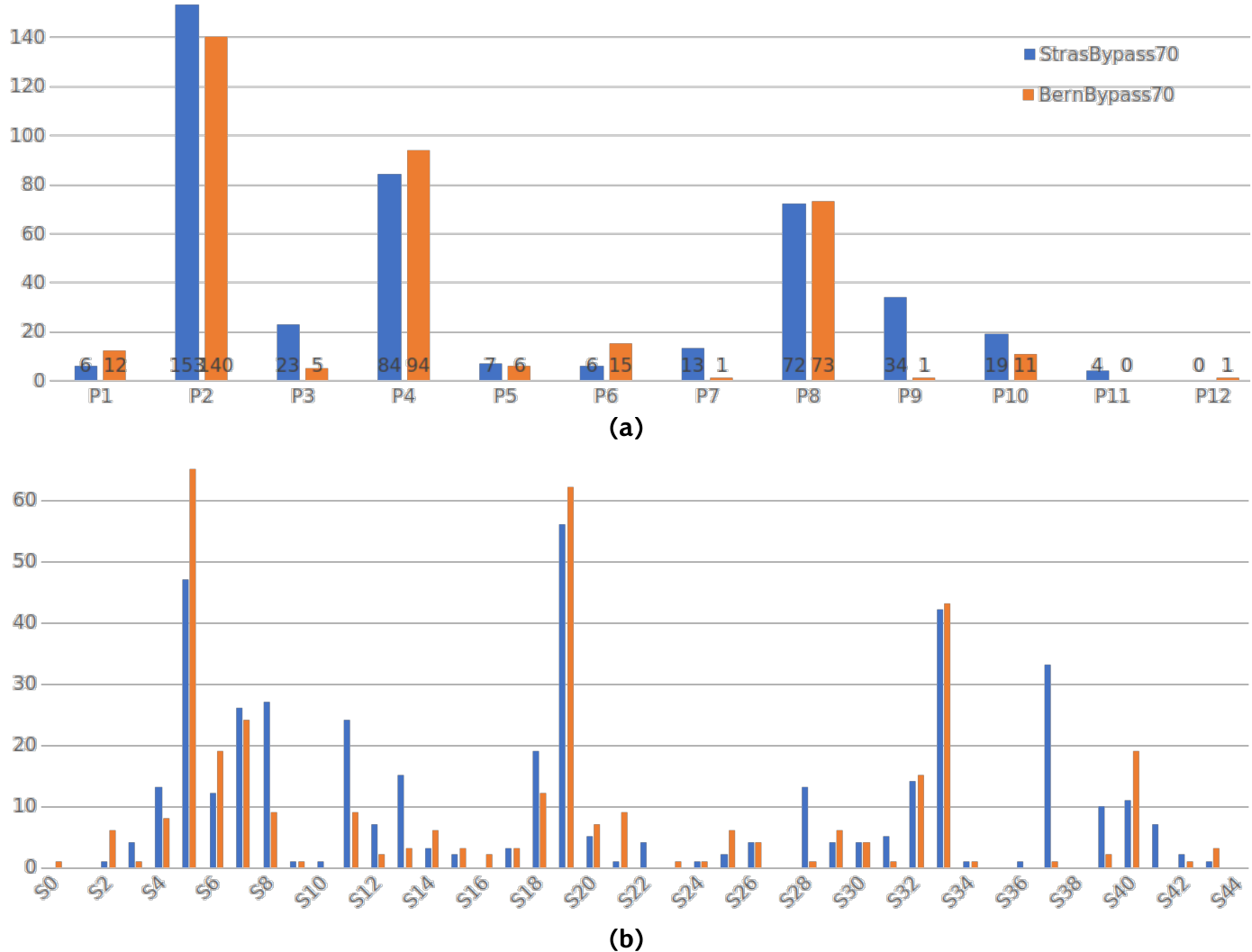}
    \caption{IAE frequency per (a) phase and (b) step across the two data centers.}
    \label{fig:correlation}
\end{figure}

We utilize the MultiBypass140 dataset \cite{lavanchy2024challenges}, which comprises 140 patient cases of Roux-en-Y gastric bypass surgery. This dataset has been extended with fine-grained annotations for IAEs, including their category and severity ~\cite{lavanchy_analyzing_2025}, see Table~\ref{tab:def}. The annotations were performed by a board-certified surgeon with over 10 years of visceral surgery experience, guided by the SEVERE index manual \cite{jung2019development} to ensure consistency and accuracy.
The dataset includes 782K frames extracted at 1 fps, with 780K frames labeled as normal and 1,594 frames annotated with IAEs (bleeding, mechanical injury, and thermal injury). Each event is labeled with start and end times, along with severity score ranging from 0 to 5, where higher values indicate greater severity. 
Sample data for each IAE type and severity level is provided in the appendix~\ref{apx:typeseverity}.
The dataset is split into 80 videos for training, 20 for validation, and 40 for testing.
Table \ref{tab:dataset} provides the distribution of IAEs across clinical centers, while Figure \ref{fig:correlation} shows the frequency of IAEs per surgical phase and step. Notably, phases 2 (gastric pouch creation), 4 (gastrojejunal anastomosis), and 8 (jejunojejunal anastomosis) are the most prevalent for IAEs. Similarly, steps 5 (lesser curvature dissection), 19 (gastrojejunal defect closure), 33 (jejunojejunal defect closure), and 37 (mesenteric defect closure) are the most affected.

\begin{table}[t]
    \centering
    \caption{Distribution of the IAEs across clinical centers present in the dataset.}
    \setlength\tabcolsep{0.09in}
    \resizebox{0.999\linewidth}{!}{%
    \begin{tabular}{@{}lcccccc@{}}
        \toprule
        Center & Cases & \#~Frames &  Normal & Bleeding & \makecell{Mechanical\\Injury} & \makecell{Thermal\\Injury}\\ \midrule
        Strasbourg & 70 & 464,973 & 426983 & 33634 & 3674 & 682\\
        Bern & 70 & 316,646 & 282204 & 28068 & 5691 & 683 \\
        \bottomrule
    \end{tabular}
    }
    \label{tab:dataset}
\end{table}

\subsection{Discrete to Continuous Severity Distribution}

\begin{figure}[!t]
    \centering
    \begin{minipage}{0.93\columnwidth}
        \centering        \includegraphics[width=1.0\textwidth]{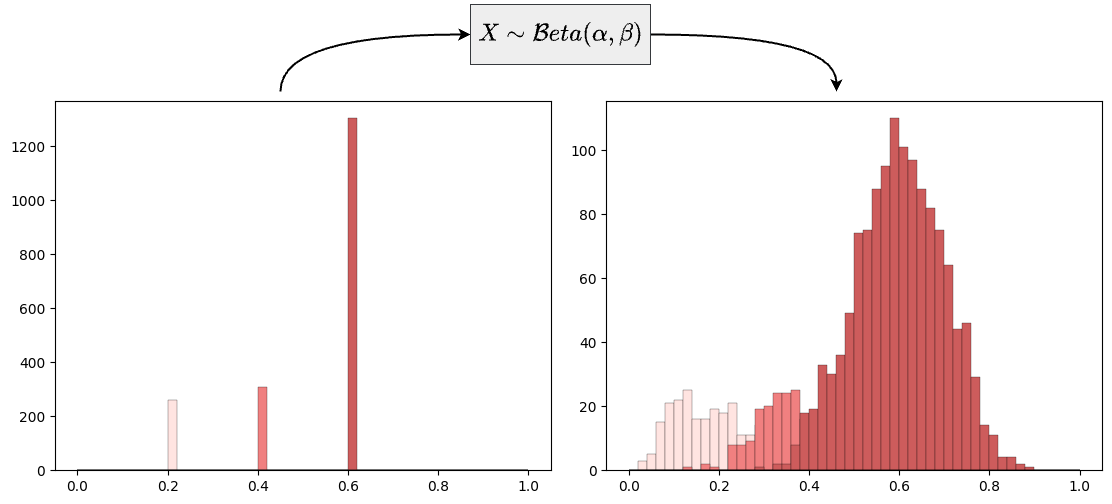} 
    \end{minipage}\hfill
    \begin{minipage}{0.93\columnwidth}
        \centering
        \includegraphics[width=1.0\textwidth]{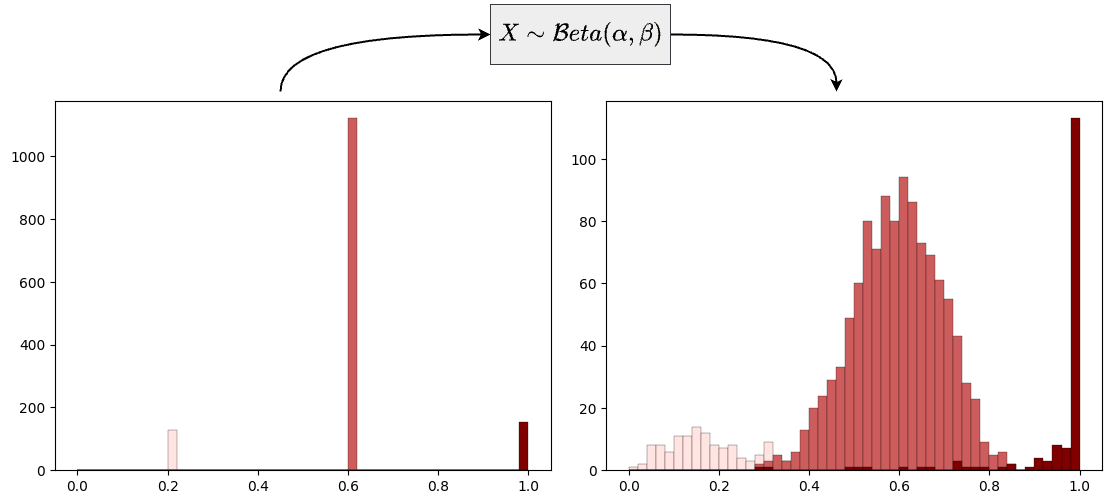} 
    \end{minipage}\hfill
    \caption{Beta distribution sampling of the grades of adverse events in (top) training and (bottom) testing sets.}
    \label{fig:databarchart}
\end{figure}

The severity levels in the dataset are annotated as discrete values, which are inherently imbalanced,  as seen in Table \ref{tab:dataset}. However, in real-world scenarios, severity exists on a continuous spectrum. To better capture this variability and address annotation noise, we propose transforming the discrete ordinal numbers into a continuous distribution using the Beta distribution as illustrated in Figure~\ref{fig:databarchart}.
The Beta distribution was chosen due to its mathematical properties and flexibility in modeling probabilistic severity scores on the normalized [0,1] interval, capturing a wide range of distributions and reflecting the inherent variability and uncertainty in clinical annotations.
The Beta distribution, defined on the interval [0, 1], is parameterized by two shape parameters, \( \alpha \) and \( \beta \), computed as:
\begin{equation}
\alpha = \mu^2 \times \left(\frac{1 - \mu}{\sigma^2} - \frac{1}{\mu}\right), \quad \beta = \alpha \times \frac{1}{\mu} - 1,
\end{equation}
where \( \mu \) represents the mean and \( \sigma \) represents the standard deviation of the distribution. These parameters enable the Beta distribution to model smooth transitions between severity levels, making it suitable for handling sparse or noisy data.
During training, we generate continuous severity values from the Beta distribution, providing a probabilistic representation that captures both the clinician's initial assessment and the inherent variability in quantification.

\begin{figure*}[t]
\centering
\includegraphics[width=0.9\textwidth]{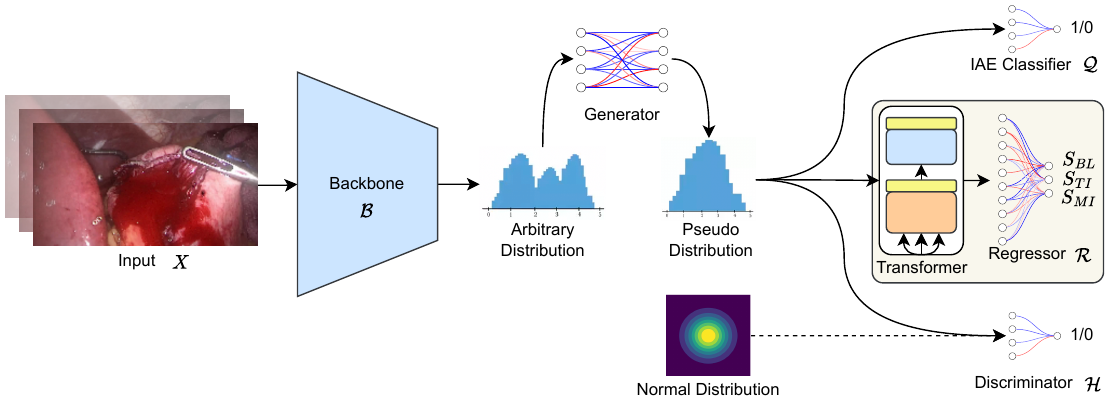}
\caption{Overview of \textit{BetaMixer}: The backbone $\mathcal{B}$ extracts features, which are transformed into a normal distribution by the generator $\mathcal{H}$. A transformer with positional embeddings encodes, classifies, and regresses IAE severity, while the discriminator $\mathcal{D}$ ensures feature normalization.}
\label{arch}
\end{figure*}

\begin{figure}[t]
\centering
\includegraphics[width=0.995\linewidth]{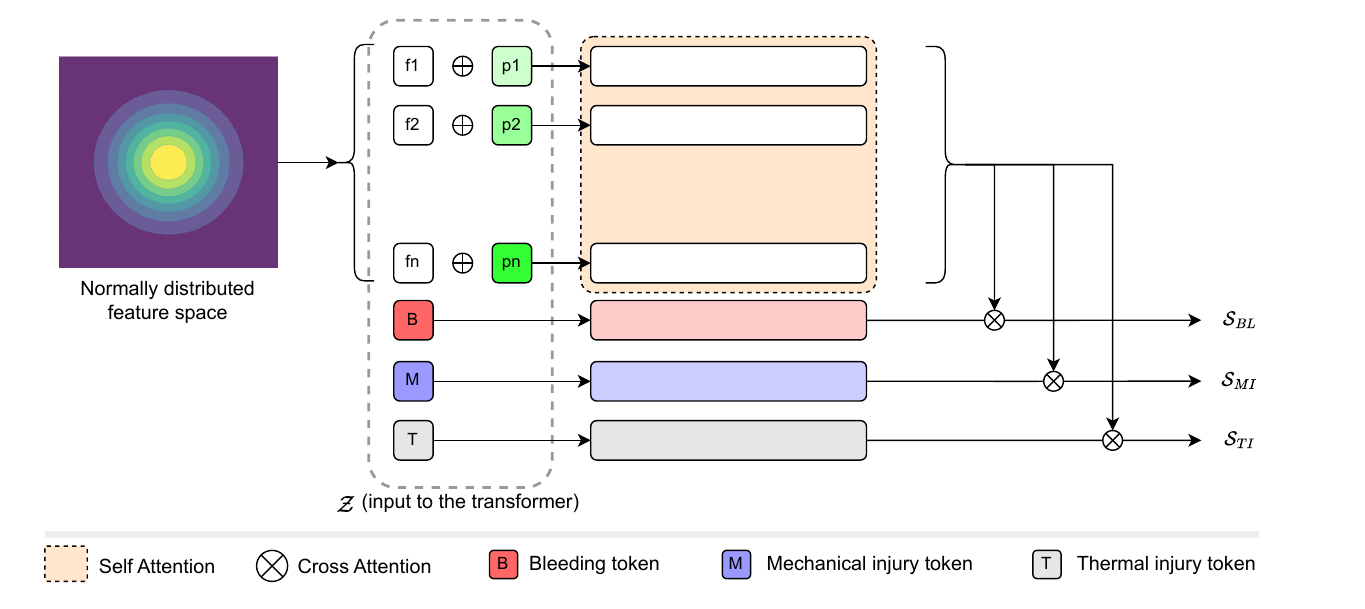}
\caption{Architecture of the transformer model - illustrating self-attention computation among frame features and independent cross-attention with the query token for a given IAE. The corresponding output is regressed to predict the severity.}\label{crossattn}
\end{figure}

\subsection{Model Architecture}
The architecture of \textit{BetaMixer} is presented in Fig. \ref{arch} and consists of
a backbone, feature generator and discriminator, IAE encoder, classifier, and regressor.

\noindent
\textbf{Backbone Feature Extractor}:  
   For each frame \( f_i \) in the input sequence \( X \), a backbone feature extractor \( \mathcal{B} \) (MobileNetV2 initialized with ImageNet weights) processes the frame and outputs a feature vector \( \hat{f_i} \in \mathbb{R}^d \), where \( d \) is the dimensionality of the feature space.

\noindent
\textbf{Normalized Feature Generator and Discriminator}:  
   A generator \( \mathcal{H} \) transforms the backbone features into normally distributed features \( \tilde{f}_i \sim \mathcal{N}(0, 1) \). The generator is implemented as a fully connected neural network (FCNN) with learnable parameters \( \theta \). A discriminator \( \mathcal{D} \) enforces that the generated features follow a standard normal distribution by classifying them as real or fake.

\noindent
\textbf{Transformer Encoder with Regression Tokens}:  
   The generated features \( \tilde{f}_i \) are passed through a transformer encoder (Fig.~\ref{crossattn}), which incorporates positional embeddings and three regression tokens \( T_{\text{BL}}, T_{\text{MI}}, T_{\text{TI}} \) for bleeding, mechanical injury, and thermal injury, respectively. The transformer applies self-attention to capture temporal dependencies across frames.

\noindent
\textbf{IAE Classifier and Regressor}:  
   The pooled features are passed into a binary classifier \( \mathcal{Q} \) for IAE detection. The regression tokens are used to predict continuous severity score for each IAE using the regression module. We use multiple regression heads to enable the detection of overlapping events with varying severity levels.


\section{Experiments}

\begin{table*}[t]
\centering
\caption{Performance and Clinical Metrics}
\label{tab:metrics}
\setlength{\tabcolsep}{8pt}
\resizebox{0.8\linewidth}{!}{%
\begin{tabular}{@{}p{1.8cm}p{5cm}p{7cm}@{}}
\toprule
\textbf{Category}           & \textbf{Metric Name}         & \textbf{Function / Equation} \\ \midrule
\textbf{Standard Metrics} & F1 Score & Balances precision and recall to evaluate the model’s performance. \newline 
   Equation: \( F1 = 2 \cdot \frac{\text{Precision} \cdot \text{Recall}}{\text{Precision} + \text{Recall}} \) \newline
   Where: \newline
   \(\text{Precision} = \frac{\text{True Positives (TP)}}{\text{TP} + \text{False Positives (FP)}}\) \newline
   \(\text{Recall} = \frac{\text{TP}}{\text{TP} + \text{False Negatives (FN)}}\). \\ \cmidrule{2-3}
 
 & Recall                       & Measures sensitivity, or the proportion of true positives identified. \newline 
   Equation: \( \text{Recall} = \frac{\text{TP}}{\text{TP} + \text{FN}} \). \\\cmidrule{2-3}
 
 & Mean Squared Error (MSE)     & Quantifies the average squared difference between predicted and actual values. \newline
   Equation: \( \text{MSE} = \frac{1}{n} \sum_{i=1}^n (y_i - \hat{y}_i)^2 \), \newline
   where \( y_i \) is the true value and \( \hat{y}_i \) is the predicted value. \\\cmidrule{2-3}
 
 & Weighted F1 Score            & Adjusts F1 score based on severity-based weights. \newline
   Equation: \( \text{Weighted F1} = \sum_{i=1}^n w_i \cdot F1_i \), \newline
   where \( w_i \) are the weights associated with severity levels. \\
                        
\midrule
                             
\textbf{Clinical Metrics}    & Classification Delay Time (CDT) & Measures time between event onset (\( T_1 \)) and correct model prediction (\( T_p \)). \newline 
   Equation: \( \text{CDT} = T_p - T_1 \). \\ \cmidrule{2-3}
 
 & Positive Predictive Value (PPV) & Evaluates the accuracy of predicting severe events (severity \( \geq 3 \)). \newline 
   Equation: \( \text{PPV} = \frac{\text{True Positives (TP)}}{\text{TP} + \text{False Positives (FP)}} \). \\ \cmidrule{2-3}
 
 & Negative Predictive Value (NPV) & Evaluates the accuracy of predicting non-severe events (severity \( \leq 1 \)). \newline 
   Equation: \( \text{NPV} = \frac{\text{True Negatives (TN)}}{\text{TN} + \text{False Negatives (FN)}} \). \\ \bottomrule
\end{tabular}}
\end{table*}

\noindent\textbf{Implementation Setup:}
We use MobileNetV2 (feature level 5) as the backbone for feature extraction. The discriminator and generator in \textit{BetaMixer} are implemented as fully convolutional networks, while the IAE classifier consists of a single convolutional layer, adaptive average pooling, and a linear layer. The transformer encoder has a projection depth of 128 with 4 transformation layers.

\noindent\textbf{Training and Loss Function:}
The model is trained adversarially for the discriminator and generator, followed by freezing the generator to train the remaining components for 30 epochs. 
Three loss functions are used: the adversarial loss trains the discriminator to distinguish between real and generated features, while the generator aims to fool the discriminator. The classification loss trains the IAE classifier to detect adverse events using binary cross-entropy. The sampled regression loss trains the transformer to predict severity scores by minimizing mean squared error (MSE) on Beta-distributed ground truth labels, which are uniformly sampled across event classes and severity levels.
Training is conducted with a batch size of 32, a spatial image resolution of \(128 \times 128\), and optimized using Adam with a learning rate of \(5 \times 10^{-5}\). All experiments are performed on a single RTX3060 GPU using the MultiBypass140 dataset.

\noindent\textbf{Baselines:}
We evaluate \textit{BetaMixer} against four baselines: two frame-based models (ResNet18 and MobileNetV2) and two temporal-based models (sMSTCN~\cite{wei2021intraoperative} and FRCNN~\cite{hua2022automatic}). ResNet18 is chosen for its deep residual structure, which effectively learns complex features, while MobileNetV2 is selected for its lightweight design, making it suitable for on-device inference. The temporal models, sMSTCN and FRCNN, have been previously explored for IAE detection tasks. All baselines are adapted to support severity regression using an extra linear layer.
We also designed an ablation model - \textit{BetaMixer (Genless)} - which the proposed model without the generator and adversarial modules.

\noindent\textbf{Evaluation Metrics:}
We evaluate \textit{BetaMixer} using standard metrics, including F1 score, recall, and Mean Squared Error (MSE), to assess performance on imbalanced datasets. The F1 score is weighted with severity-based weights \([0.02, 0.06, 0.12, 0.19, 0.26, 0.33]\) to balance sensitivity and specificity. For continuous severity prediction, thresholds of \(0.5\) (classification) and \((0.2, 0.4, 0.6, 0.8)\) (regression) are applied.
Additionally, we use clinically relevant metrics: Classification Delay Time (CDT), Positive Predictive Value (PPV), and Negative Predictive Value (NPV). CDT measures the delay between the first occurrence of an event and its correct prediction. PPV evaluates the accuracy of predicting severe events (levels \(\geq 3\)), while NPV assesses the prediction of non-severe events (levels \(\leq 1\)), ensuring minimal unnecessary interventions.
Further information on the metrics can be found in the Table~\ref{tab:metrics}.

\section{Results and Discussion}\label{sec:results}

\begin{table}[!t]
    \caption{Mean IAE classification (F1/recall) and regression (MSE) results of BetaMixer in comparison with baselines.}
    \label{tab:recall}
    \setlength{\tabcolsep}{15pt}
    \resizebox{\linewidth}{!}{
    \begin{tabular}{@{}lccc@{}} \toprule
    Model & F1 $\uparrow$ & Recall $\uparrow$ & MSE $\downarrow$ \\ \midrule
    ResNet18 & 0.68$\pm$0.12 & 0.76$\pm$0.15 & 0.30$\pm$0.14 \\
    MobileNet & 0.68$\pm$0.18 & 0.74$\pm$0.13 & 0.32$\pm$0.20 \\
    sMSTCN & 0.71$\pm$0.14 & 0.79$\pm$0.12 & 0.28$\pm$0.16 \\
    FRCNN  & 0.72$\pm$0.15 & 0.78$\pm$0.16 & 0.26$\pm$0.19 \\ \hline
    BetaMixer (Ours)   & \textbf{0.76$\pm$0.12} & \textbf{0.81$\pm$0.14} & \textbf{0.23$\pm$0.15 } \\ \bottomrule
    \end{tabular}
    }
\end{table}

\begin{table*}[!t]
    \centering
    \caption{Results of IAE classification on 5 seconds clip over the whole testing set.}
    \label{tab:classification}    
    \setlength{\tabcolsep}{8pt}
    \resizebox{0.98\linewidth}{!}{%
    \begin{tabular}{@{}lcccrcccrcccrccc@{}}
        \toprule
        \multirow{2}{*}{Model} & \multicolumn{3}{c}{Bleeding} & \phantom{abc} & \multicolumn{3}{c}{Mechanical injury} & \phantom{abc} & \multicolumn{3}{c}{Thermal injury} & \phantom{abc} & \multicolumn{3}{c}{Overall IAE} \\
         \cmidrule{2-4}
        \cmidrule{6-8}
        \cmidrule{10-12}
        \cmidrule{14-16}      
        & F1 & PPV & NPV && F1 & PPV & NPV && F1 & PPV & NPV && F1 & PPV & NPV \\
        \midrule
        ResNet18 & 0.72&0.71&0.77 && 0.63&0.62&0.59 && 0.70&0.72&0.66 && 0.68&0.66&0.67\\
        MobileNetV2 & 0.71&0.72&0.76 && 0.61&0.66&0.61 && 0.72&0.70&0.71 && 0.68&0.68&0.69\\
        sMSTCN\cite{wei2021intraoperative} & 0.75&0.71&0.77 && 0.64&0.65&0.61 && 0.75&0.71&\bf0.79 && 0.71&0.69&0.72 \\
        FRCNN\cite{hua2022automatic} & 0.76&0.73&0.78 && 0.68&0.67&0.60 && 0.74&0.72&0.77 && 0.72&0.70&0.71\\ \midrule
        BetaMixer (Genless)   & 0.72&0.71&0.76 && 0.65&0.66&0.59 && 0.72&0.71&0.73 && 0.69&0.69&0.70\\
        BetaMixer (Proposed)   & \bf0.81&\bf0.76&\bf0.82 && \bf0.70&\bf0.69&\bf0.63 && \bf0.77&\bf0.74&0.77 && \bf0.76&\bf0.73&\bf0.84\\
        
         \bottomrule
    \end{tabular}
    }
\end{table*}

\begin{table*}[!t]
    \centering
    \caption{Results of IAE regression for k = 5 different severity grades in terms of mean squared error.}
    \label{tab:regression}    
    \setlength{\tabcolsep}{10pt}
    \resizebox{0.98\linewidth}{!}{%
    \begin{tabular}{@{}lcccccccccccccc@{}}
        \toprule
        \multirow{2}{*}{Model} & \multicolumn{5}{c}{Bleeding} & \phantom{abc} & \multicolumn{3}{c}{Mechanical injury} & \phantom{abc} & \multicolumn{4}{c}{Thermal injury} \\
         \cmidrule{2-6}
        \cmidrule{8-10}
        \cmidrule{12-15}      
        & 0 & 1 & 2 & 3 & 4 && 0 & 1 & 3 && 0 & 1 & 3 & 5 \\
        \midrule
        ResNet18 & 0.47&0.26&0.31&0.28&0.29 && 0.27&0.19&0.27 && 0.25&0.20&0.20&0.69\\
        MobileNet & 0.51&0.27&0.31&0.29&0.30 && 0.29&0.23&0.29 && 0.27&0.23&0.21&0.68\\
        sMSTCN\cite{wei2021intraoperative} & 0.35&0.25&0.32&0.30&0.29 && 0.21&0.22&0.20 && 0.23&0.17&0.23&0.63\\
        FRCNN\cite{hua2022automatic} & 0.35&0.22&0.30&0.28&0.27 && 0.19&0.15&0.21 && 0.19&0.18&0.23&0.65\\ \midrule
        BetaMixer (Genless) & 0.34&0.23&0.30&0.29&0.28&&0.22&0.14&0.21&&0.19&0.17&0.24&0.61\\
        BetaMixer (Proposed) & 0.30&0.21&0.26&0.24&0.25 && 0.19&0.10&0.18 && 0.17&0.14&0.19&0.57\\
        
         \bottomrule
    \end{tabular}
    }
\end{table*}

\begin{figure*}[!t]
\centering
\includegraphics[width=0.95\linewidth]{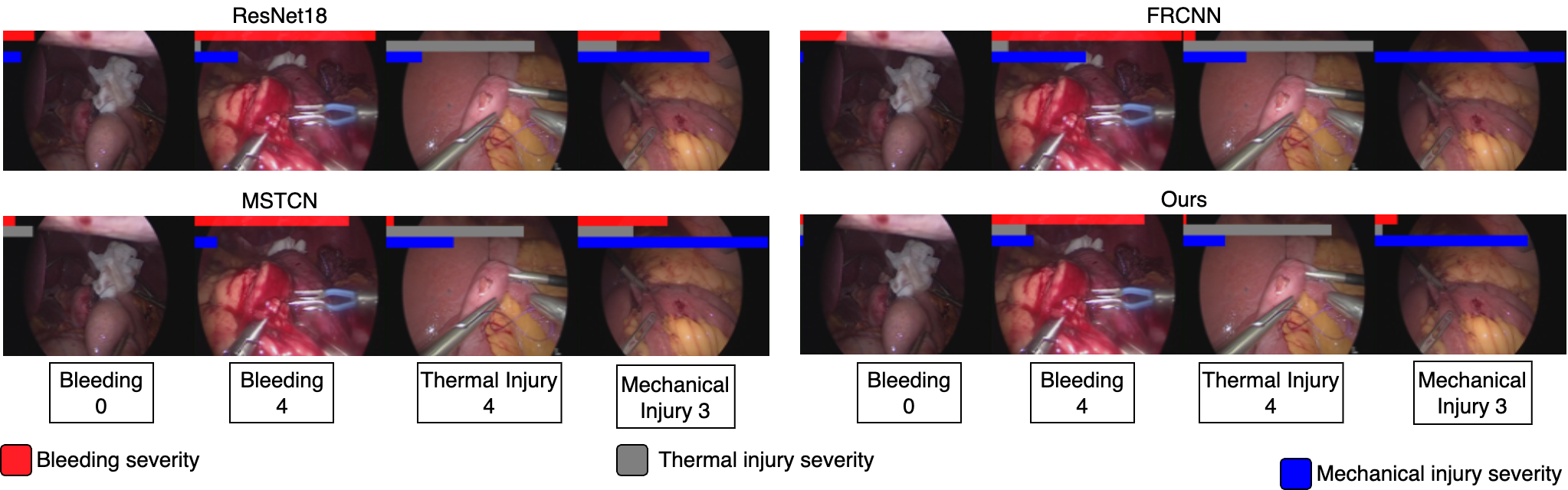}
\caption{The visual performance of our model compared with the baselines. The length of the bar indicates the severity predicted by the respective model. The groundtruth are mentioned in the box for a column.}
\label{fig:qualitative}
\end{figure*}

Our proposed \textit{BetaMixer} model demonstrates superior performance in IAE classification and severity regression compared to existing baselines, as shown in Tables \ref{tab:recall}. Overall, \textit{BetaMixer} achieves a $+4\%$ improvement in F1 score, $+3\%$ in recall, $+3\%$ in PPV, and $+13\%$ in NPV. Notably, it excels in detecting and quantifying bleeding and mechanical injury, achieving the best scores across all metrics as seen in Table~\ref{tab:classification}. For thermal injury, while the NPV is slightly lower ($-2\%$) compared to FRCNN, \textit{BetaMixer} outperforms in PPV and F1 score, likely due to the presence of smoke from coagulation tools, which serves as a strong visual indicator for this IAE.

In terms of severity regression (Table \ref{tab:regression}), \textit{BetaMixer} consistently performs well across all IAE categories. It achieves the lowest mean squared error (MSE) of 0.2 for level 1 bleeding and 0.1 for level 1 mechanical injury. However, predicting higher-severity thermal injury (level 5) remains challenging due to the absence of such samples in the training set. This highlights the need for more diverse training data to improve performance on rare, high-severity events.

The Classification Delay Time (CDT) metric, evaluated in Table \ref{tab:delay}, further underscores the effectiveness of \textit{BetaMixer}. It achieves the lowest CDT for bleeding (1.31) and mechanical injury (1.12), outperforming sMSTCN and FRCNN. For thermal injury, the CDT is slightly higher, indicating room for improvement in detecting this specific IAE. These results demonstrate the model's ability to provide timely predictions, which is critical for intraoperative decision-making.

\begin{table}[!t]
    \caption{Result of Classification delay time (in secs) of IAE on the test using 5 seconds clip window.}
    \label{tab:delay}
    \setlength{\tabcolsep}{6pt}
    \resizebox{\linewidth}{!}{
    \begin{tabular}{@{}lcccc@{}} \toprule
    Model & Bleeding & Mech. injury & Thermal injury & Mean\\ \midrule
    ResNet18 & 1.53 & 1.51 & 1.23 & 1.42 \\
    MobileNet & 1.40 & 1.41 & 1.21 & 1.34 \\
    sMSTCN & 1.43 & 1.42 & 1.10 & 1.31 \\
    FRCNN  & 1.41 & 1.23 & \textbf{0.91} & 1.27 \\ \hline
    BetaMixer (Ours)  & \textbf{1.31} & \textbf{1.12} & {1.13} & \textbf{1.23}    \\ \bottomrule
    \end{tabular}
    }
\end{table}

\begin{table*}[!t]
    \centering
    \caption{Ablation on the performance of BetaMixer model on the length of clips.}
    \label{tab:ablation}    
    \setlength{\tabcolsep}{10pt}
    \resizebox{0.98\linewidth}{!}{%
    \begin{tabular}{@{}lcccrcccrcccrccc@{}}
        \toprule
        \multirow{2}{*}{Frames} & \multicolumn{3}{c}{Bleeding} & \phantom{abc} & \multicolumn{3}{c}{Mechanical injury} & \phantom{abc} & \multicolumn{3}{c}{Thermal injury} & \phantom{abc} & \multicolumn{3}{c}{Overall IAE} \\
         \cmidrule{2-4}
        \cmidrule{6-8}
        \cmidrule{10-12}
        \cmidrule{14-16}      
        & F1 & PPV & NPV && F1 & PPV & NPV && F1 & PPV & NPV && F1 & PPV & NPV \\
        \midrule
        1 & 0.78&0.72&0.79 && 0.65&0.67&0.60 && 0.71&0.71&0.70 && 0.71&0.70&0.69\\
        5 & \bf0.81&\bf0.76&\bf0.82 && \bf0.70&\bf0.69&\bf0.63 && \bf0.77&\bf0.74&\bf0.77 && \bf0.76&\bf0.73&\bf0.84 \\
        10 & 0.77&0.74&0.80 && 0.68&0.66&0.61 && 0.73&0.70&0.74 && 0.73&0.70&0.74 \\
        25 & 0.75&0.75&0.80 && 0.67&0.66&0.60 && 0.74&0.69&0.73 && 0.72&0.70&0.71\\ 
        
         \bottomrule
    \end{tabular}
    }
\end{table*}

An ablation study on the impact of input sequence length (Table \ref{tab:ablation}) reveals that a 5-frame input yields the best performance across all IAE categories, with an F1 score of 0.76, PPV of 0.73, and NPV of 0.84. This suggests that IAEs are temporal events best captured within short intervals, as performance degrades with longer or shorter sequences. 
Evaluating \textit{BetaMixer} without the generator component (\textit{BetaMixer (Genless)} in Table \ref{tab:regression}) reveals a performance drop, underscoring the generator's role in feature normalization and overall accuracy improvement.


Qualitative results in Fig. \ref{fig:qualitative} further validate that \textit{BetaMixer} more accurately approximates ground truth in both classification and severity regression compared to baselines. From these observations, \textit{BetaMixer} sets a new benchmark for IAE detection and severity regression, demonstrating robustness in handling imbalanced datasets and providing timely, accurate predictions. Its performance underscores the importance of temporal modeling and continuous severity representation in surgical AI systems.

\section{Conclusions}
This paper addresses the challenge of classifying and quantifying intraoperative adverse events (IAEs), such as bleeding, thermal injury, and mechanical injury, during laparoscopic gastric bypass surgery. To tackle data imbalance caused by the rarity of these events, we propose \textit{BetaMixer}, a novel approach integrating normalized feature mixing, Beta distribution-based sampling, and continuous feature space regularization. Our method outperforms baselines in IAE classification and severity regression, achieving superior performance across automated and clinical metrics. An ablation study reveals optimal performance with 5-frame inputs, emphasizing the importance of temporal modeling for short-duration events. \textit{BetaMixer} quantifies adverse events on a 0 to 5 scale with high accuracy, mitigating data imbalance and providing a robust solution for real-time IAE detection and severity assessment. While this work focused on Roux-en-Y gastric bypass due to data availability, future work will explore generalizability to other surgical domains and incorporate additional data sources to enhance performance.
To facilitate continued research in this area and direction, the IAE annotations is integrated into the existing \mbox{MultiBypass140} dataset~\cite{lavanchy2024challenges}, which is publicly released and accessible via \mbox{\url{https://github.com/CAMMA-public/MultiBypass140}.}


\section*{Acknowledgements} 
This work was supported by French state funds managed within the Plan Investissements d’Avenir by the ANR under references: National AI Chair AI4ORSafety [ANR-20-CHIA- 0029-01], IHU Strasbourg [ANR-10-IAHU-02] and by BPI France [Project 5G-OR]. Joel L. Lavanchy acknowledges funding by the Swiss National Science Foundation (P500PM206724, P5R5PM217663). 
This work has also received funding from the European Union (ERC, CompSURG, 101088553). Views and opinions expressed are however those of the authors only and do not necessarily reflect those of the European Union or the European Research Council. Neither the European Union nor the granting authority can be held responsible for them.


\bibliographystyle{IEEEtran}
\bibliography{arxiv}

\onecolumn

\appendix
\section{Appendix}\label{apx:typeseverity}

\begin{figure*}[!thbp]
     \centering
    \includegraphics[width=0.995\linewidth]{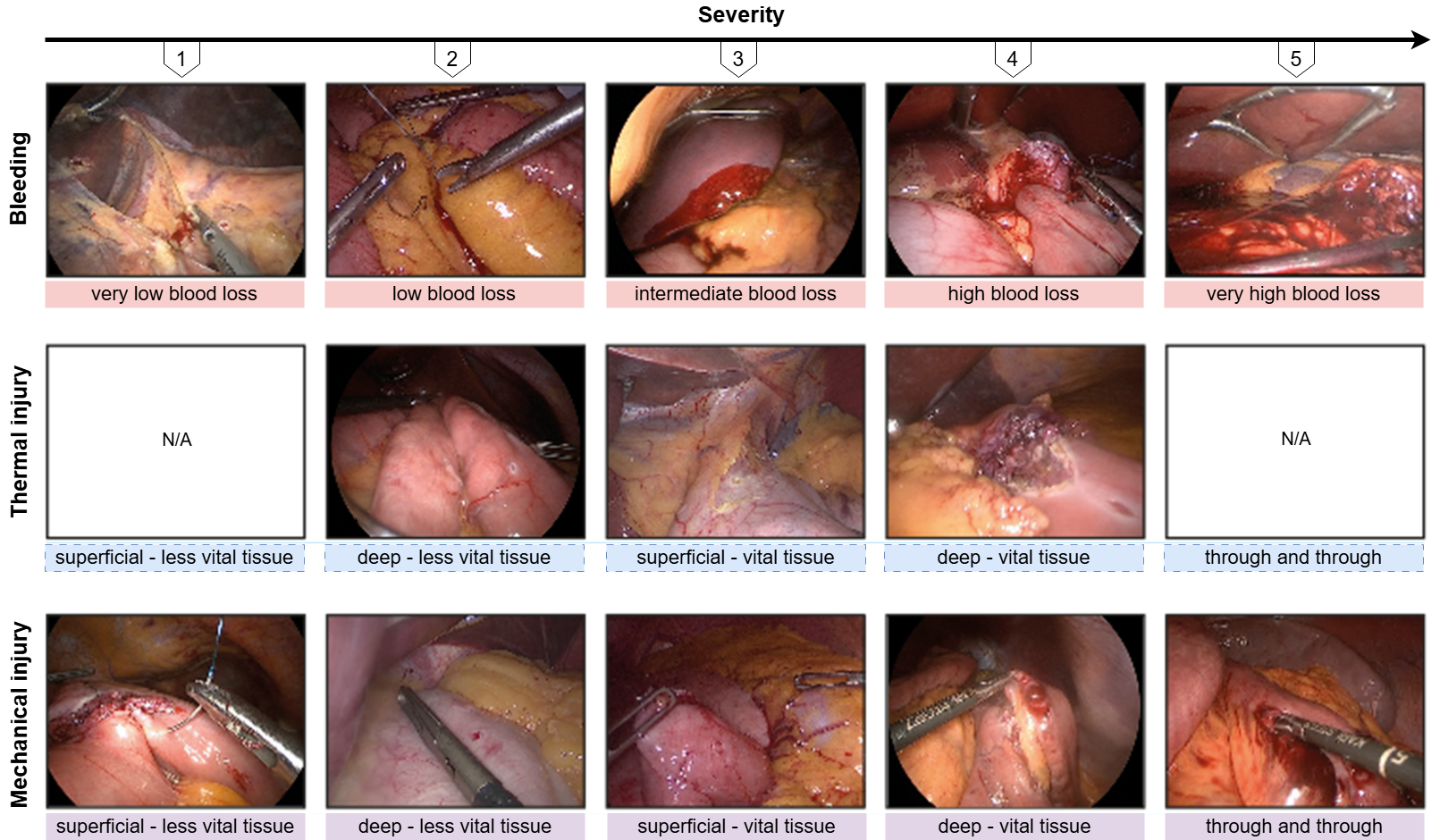}
    \caption{Sample images and severity levels in the MultiBypass140 dataset.}
    \label{fig:typeseverity}
\end{figure*}

\end{document}